# Tightening Fractional Covering Upper Bounds on the Partition Function for High-Order Region Graphs

**Tamir Hazan**
TTI Chicago
Chicago, IL 60637

**Jian Peng**
TTI Chicago
Chicago, IL 60637

**Amnon Shashua**
The Hebrew University
Jerusalem, Israel

## Abstract

In this paper we present a new approach for tightening upper bounds on the partition function. Our upper bounds are based on fractional covering bounds on the entropy function, and result in a concave program to compute these bounds and a convex program to tighten them. To solve these programs effectively for general region graphs we utilize the entropy barrier method, thus decomposing the original programs by their dual programs and solve them with dual block optimization scheme. The entropy barrier method provides an elegant framework to generalize the message-passing scheme to high-order region graph, as well as to solve the block dual steps in closed-form. This is a key for computational relevancy for large problems with thousands of regions.

## 1 Introduction

A large set of reasoning problems can be framed as a set of dependency relations among possible structures. These dependencies, usually expressed by a graph, define a joint probability function which drives an inference engine over those structures [18]. Such graphical models, also known as Markov Random Fields (MRFs), are found in a wide variety of fields and applications, including object detection [5], stereo vision [30], parsing [19], or protein design [29], as well as other broad disciplines which include artificial intelligence, signal processing and statistical physics.

The inference problem in MRFs involves assessing the likelihood of possible structures, whether objects, parsers, or molecular structures. The structures are specified by assignments of random variables, whose scores are described in a concise manner by interactions over small subsets of variables. In a fully probabilistic treatment, all possible alternative assignments are considered. However, this requires summing over the assignments with their respective weights – evaluating the partition function. The partition function has a special role which goes beyond that of assigning a probability to the alternative assignments. In addition it plays a fundamental role in various contexts including approximate inference [26], maximum-likelihood parameter estimation [20] and large deviation bounds [4]. Computing the partition function requires summing over the assignments with their respective weights, thus it is a #P hard problem (i.e., efficient weighted counting is unlikely achievable) [31]. Instead there is much focus on finding approximate solutions and bounds [34].

In this paper we propose an iteration of concave and convex programs, each using a primal-dual iterative scheme, to compute increasingly tight upper bounds on the partition function that are based on fractional covering bounds on the entropy [6, 23]. Our work is distinct on a couple of fronts: we handle general region graphical models, unlike previous attempts that focus on particular graphs like bipartite forms (outer and inner regions) or on limited size factors (like pairwise). Secondly, our method is based on achieving a message-passing constellation between nodes of the region graph thereby utilizing the local structure of the problem. A message-passing framework, or more generally the dual decomposition framework, is key for computational relevancy for large problems which contain scores of thousands of regions where the factors are defined on a relatively small number of variables (e.g., when variables correspond to a large grid of points).

We begin by introducing the notation and the fractional covering upper bounds of interest. We subsequently present a message-passing algorithm to compute the upper bound, using dual decomposition. We also introduce a new approach to tighten the fractional covering upper bounds based on the entropy barrier function and dual block optimization, and demonstrate the effectiveness of the approach.

## 2 Notations, Problem Setup and Background

Let $x = (x_1, ..., x_n)$ be the realizations of $n$ discrete random variables where the range of the $i'th$ random variable is $\{1, ..., n_i\}$, i.e., $x_i \in \{1, ..., n_i\}$. We consider a joint distribution $p(x_1, ..., x_n)$ and assume that it factors into a product of non-negative functions $\psi_r(x_r)$, known as *potentials*. Usually, the potentials are defined over a small subset of indexes $r \subset \{1, ..., n\}$, called *regions*:

$$p(x_1, ..., x_n) = \frac{1}{Z} \prod_{r \in \mathcal{R}} \psi_r(x_r)$$

where for singletons, i.e. $|r| = 1$, the functions $\psi_r(x_r)$ represent "local evidence" or prior data on the states of $x_i$, and for $|r| > 1$ the potential functions describe the interactions of their variables $x_r \subset \{x_1, ..., x_n\}$, and $Z$ is the normalization constant, also called the *partition function*. For convenience, we adopt the additive form by setting $\phi_r(x_r) = \ln \psi_r(x_r)$ thereby having the joint probability take the form of a Gibbs distribution:

$$p(x_1, , x_n) = e^{\sum_{r \in \mathcal{R}} \phi_r(x_r) - \ln Z}$$

For example, $p(x_1, x_2, x_3) \propto \exp(\phi_2(x_2) + \phi_{123}(x_1, x_2, x_3) + \phi_{23}(x_2, x_3))$ has three factors with regions $\mathcal{R} = \{2\}, \{1, 2, 3\}, \{2, 3\}$. The factorization structure above defines a hypergraph whose nodes represent the $n$ random variables, and the regions $r \in \mathcal{R}$ correspond to its hyperedges. A convenient way to represent this hypergraph is by a *region graph*. A region graph is a directed graph whose nodes represent the regions and its direct edges correspond to the inclusion relation, i.e., a directed edge from node $r$ to $s$ is possible only if $s \subset r$. We adopt the terminology where $P(r)$ and $C(r)$ stand for all nodes that are parents and children of the node $r$, respectively. Also, we define $R_i$ to be the set of regions which contains the variable $i$.

Inference is closely coupled with the ability to evaluate the logarithm of the partition function $\ln Z$. From a variational perspective, there is a relationship between the (minus) Gibbs-Helmholtz free-energy and $\ln Z$:

$$\ln Z = \max_{\substack{p(x) \geq 0, \\ \sum_x p(x) = 1}} \left\{ \sum_{r \in \mathcal{R}} \sum_{x_r} p(x_r) \phi_r(x_r) + H(p) \right\} \quad (1)$$

where $p(x_r) = \sum_{x \setminus x_r} p(x)$ is the marginal probability and $H(p) = -\sum_x p(x) \ln p(x)$ is the entropy of the distribution. However, the complexity of the variational representation is unwieldy because both the entropy and the simplex constraint require an evaluation over all possible states of the system $x = (x_1, ..., x_n)$, which is exponential in $n$. Instead one looks for an approximation or bounds. An upper bound is designed with a tractable approximation of the free-energy by (i) upper bounding the entropy term $H(p)$ by a combination of local entropies over marginal probabilities $p(x_r)$, and (ii) by outer bounding the probability simplex constraints by the so called "local consistency" constraints.

An upper bound on the entropy function $H(p)$ proceeds by replacing the entropy by the fractional covering entropy bounds [6, 23]. These upper bounds are defined as sum of local entropies over the marginal probabilities $H(p(x_r)) = -\sum_{x_r} p(x_r) \ln p(x_r)$:

$$H(p) \leq \sum_{r \in \mathcal{R}} c_r H(p(x_r))$$

These upper bounds hold whenever $c_r \geq 0$ and for every $i = 1, ..., n$ there holds $\sum_{r \in R_i} c_r = 1$, where $R_i$ is the set of all regions that contain $i$. The second step in obtaining an efficient upper bound is replacing the marginal distributions $p(x_r)$ by "beliefs" $b_r(x_r)$. The beliefs form "pseudo distributions" in the sense that the beliefs might not necessarily arise as marginal probabilities of some distribution $p(x_1, ..., x_n)$. To maintain local consistency between beliefs which share the same variables, we define the *local consistency polytope* $L(G)$ for the region graph $G$, as follows:

$$L(G) = \left\{ \{b_r\}_{r \in \mathcal{R}} : \begin{array}{l} \sum_{x_s \setminus x_r} b_s(x_s) = b_r(x_r) \quad \forall r, s \in P(r) \\ b_r \geq 0, \sum_{x_r} b_r(x_r) = 1 \quad \forall r \in \mathcal{R} \end{array} \right.$$

For example, assume $\mathcal{R}$ consists of the three factors $\{1\}, \{1, 2\}, \{1, 3\}$ then the consistency constraints on the beliefs $b_1(x_1), b_{1,2}(x_1, x_2), b_{1,3}(x_1, x_3)$ enforce their distribution constraints and marginalization constraints $\sum_{x_3} b_{1,3}(x_1, x_3) = b_1(x_1)$ and $\sum_{x_2} b_{1,2}(x_1, x_2) = b_1(x_1)$.

Taken together, the upper bound on the log-partition is defined as follows:

$$\ln Z \leq \max_{b_r(x_r) \in L(G)} \left\{ \sum_{r, x_r} b_r(x_r) \phi_r(x_r) + \sum_r c_r H(b_r) \right\} \quad (2)$$

Thus we introduce a family of upper bounds for the partition function, for the set of fractional covering numbers $c_r$. The computational complexity of these upper bounds is no longer exponential in $n$, but linear in the number of the regions and the number of assignments in each region.

**Computing the fractional covering upper bound:** Given potential functions $\phi_r(x_r)$, and nonnegative covering numbers $c_r$. Set $c_{p,r} = c_p/(c_r + \sum_{p' \in P(r)} c_{p'})$. for every initial values of messages $\lambda_{c \to p}(x_c)$:

1. For $t = 1, 2, ...$

   (a) For $r \in \mathcal{R}$ do:

   $$\forall x_r, \forall p \in P(r) \quad \mu_{p \to r}(x_r) = c_p \log \Big( \sum_{x_p \backslash x_r} \exp \Big( (\phi_p(x_p) + \sum_{c \in C(p) \backslash r} \lambda_{c \to p}(x_c) - \sum_{p' \in P(p)} \lambda_{p \to p'}(x_p)) \Big/ c_p \Big) \Big)$$

   $$\forall x_r, \forall p \in P(r) \quad \lambda_{r \to p}(x_r) = c_{p,r} \Big( \phi_r(x_r) + \sum_{c \in C(r)} \lambda_{c \to r}(x_c) + \sum_{p' \in P(r)} \mu_{p' \to r}(x_r) \Big) - \mu_{p \to r}(x_r)$$

2. Output:

   (beliefs) $\quad \forall r \in \mathcal{R} \quad c_r \neq 0 : b_r^*(x_r) \propto \exp \Big( (\phi_r(x_r) + \sum_{c \in C(r)} \lambda_{c \to r}(x_c) - \sum_{p \in P(r)} \lambda_{r \to p}(x_r)) \Big/ c_r \Big)$

   $\quad \forall r \in \mathcal{R} \quad c_r = 0 : support(b_r^*) \subset argmax_{x_r}\{\phi_r(x_r) + \sum_{c \in C(r)} \lambda_{c \to r}(x_c) - \sum_{p \in P(r)} \lambda_{r \to p}(x_r)\}$

   (bound) $\quad \sum_{r, x_r} b_r^*(x_r) \phi_r(x_r) + \sum_r c_r H(b_r^*)$

Figure 1: *The fractional covering upper bound appears in equation (2). The support of the beliefs are their non-zero entries, and when $c_r = 0$ it corresponds to the max-beliefs. When considering bipartite region graphs, this algorithm reduces to many of the previous message-passing algorithms, see Section 6.*

## 3 Computing High-Order Upper Bounds

In the following we develop an efficient message-passing method to compute the region based upper bounds and their optimal beliefs $b_r(x_r)$, for fixed value of covering numbers $c_r$, as described in equation (2). These upper bounds depend on the non-negative fractional covering numbers, therefore correspond to maximizing a concave function subject to convex constraints. Such concave programs can be solved by minimizing their dual convex programs. Nevertheless, there are potentially many different convex dual programs, depending on the set of constraints, or Lagrange multipliers, one aims at satisfying. We realize that the probability simplex constraints for $b_r(x_r)$ are easier to satisfy, therefore we derive a dual program which ignores these constraints. For this purpose we use the entropy function as a barrier function over the probability simplex.

**Theorem 1.** *Define the entropy as a barrier function over the probability simplex,*

$$H(b_r) = \begin{cases} -\sum_{x_r} b_r(x_r) \log b_r(x_r) & \text{if } b_r \in \Delta \\ -\infty & \text{otherwise} \end{cases}$$

*where $b_r \in \Delta$ if $b_r(x_r) \geq 0$ and $\sum_{x_r} b_r(x_r) = 1$. The fractional covering numbers in equation (2) are non-negative, thus the bound is a concave function and its dual function takes the form*

$$D(\lambda) = \sum_r c_r \log \Big( \sum_{x_r} \exp(\hat{\phi}_r(x_r)/c_r) \Big)$$

*where*

$$\hat{\phi}_r(x_r) = \phi_r(x_r) + \sum_{c \in C(r)} \lambda_{c \to r}(x_c) - \sum_{p \in P(r)} \lambda_{r \to p}(x_r)$$

*In particular, strong duality holds and the primal optimal solution can be derived from the dual optimal solution*

$$b_r(x_r) \propto \exp \Big( \hat{\phi}_r(x_r) \Big/ c_r \Big)$$

*Whenever $c_r = 0$ the corresponding primal optimal solution $b_r(x_r)$ corresponds to the max-beliefs, i.e., probability distributions over the maximal arguments of $\hat{\phi}_r(x_r)$.*

**Proof:** Since we use the entropy as a barrier function over the probability simplex we only need to apply Lagrange multipliers $\lambda_{r \to p}(x_r)$ for the marginalization constraints $b_r(x_r) = \sum_{x_p \backslash x_r} b_p(x_p)$ for every region $r \in \mathcal{R}$, every assignment $x_r$ and every parent $p \in P(r)$. Therefore the Lagrangian takes the form

$$L(b_r, \lambda_{c \to p}) = \sum_{r, x_r} b_r(x_r) \phi_r(x_r) + \sum_r c_r H(b_r)$$

$$+ \sum_{r, x_r, p \in P(r)} \lambda_{r \to p}(x_r) \Big( \sum_{x_p \backslash x_r} b_p(x_p) - b_r(x_r) \Big)$$

The dual function is recovered by maximizing the beliefs $b_r(x_r)$ in the Lagrangian. Thus

$$D(\lambda) = \sum_r \max_{b_r} \{\sum_{x_r} b_r(x_r)\hat{\phi}_r(x_r) + c_r H(b_r)\}$$
$$= \sum_r c_r \max_{b_r} \{\sum_{x_r} b_r(x_r)(\hat{\phi}_r(x_r)/c_r) + H(b_r)\}$$

The maximization over the beliefs in the Lagrangian is done while satisfying the probability simplex constraint as they are encoded in the domain of the entropy function. The result follows from the duality between the entropy barrier function over the probability simplex and the log-partition function, c.f. equation (1). Strong duality holds using Theorem 6.2.5 in [1]. The primal optimal solution is derived from the dual optimal solution as the primal arguments that maximize the Lagrangian. □

The theorem above uses the conjugate duality between the entropy barrier function, weighted by the non-negative number $c_r$, and the weighted extension of the log-partition. The weighted log-partition is called the soft-max function and it is used as a smooth approximation for the max function whenever $c_r \to 0$. In particular, when $c_r = 0$ the soft-max function reduces to the max-function, and Theorem 1 demonstrates the known conjugate duality between the indicator function over the probability simplex and the max-function.

One of the most important properties of the dual function is that it decomposes the constraints set, a feature that is typically called dual decomposition. In Theorem 1 every dual variable $\lambda_{c \to p}(x_c)$ represents a marginalization constraint $\sum_{x_p \backslash x_c} b_p(x_p) = b_c(x_c)$ in the primal. Therefore optimizing a single dual variable amounts to solving the primal problem with a single constraint. Therefore, the dual coordinate descent algorithm decomposes the primal problem complexity to smaller sub-problems, while the dual function encodes the consistency between the sub-problems solutions. Moreover, since the marginalization constraints encode the graphical model, performing block coordinate descent results in sending messages along the edges of the region graph, thus we are able to cope with large-scale and high-order graphical models.

**Theorem 2.** *Block coordinate descent on the dual in Theorem 1 takes the form: For every region $r \in \mathcal{R}$*

$\forall x_r, \forall p \in P(r)$

$$\mu_{p \to r}(x_r) = c_p \log \Big( \sum_{x_p \backslash x_r} \exp(\phi_{p,r}(x_p)/c_p) \Big)$$

$$\lambda_{r \to p}(x_r) = c_{p,r} \Big( \phi_r(x_r) + \sum_{c \in C(r)} \lambda_{c \to r}(x_c) + \sum_{p' \in P(r)} \mu_{p' \to r}(x_r) \Big)$$
$$\quad - \mu_{p \to r}(x_r)$$

*where $c_{p,r} = c_p/(c_r + \sum_{p' \in P(r)} c_{p'})$ and $\phi_{p,r}(x_p) = \phi_p(x_p) + \sum_{c \in C(p) \backslash r} \lambda_{c \to p}(x_c) - \sum_{p' \in P(p)} \lambda_{p \to p'}(x_p)$. If a region covering number in the dual objective equals zero, i.e., $c_r = 0$, then the block coordinate descent update rules for this region hold for every non-negative $c_r$.*

**Proof:** The gradient equals to

$$\frac{\partial D}{\partial \lambda_{r \to p}(x_r)} = \sum_{x_p \backslash x_r} b_p(x_p) - b_r(x_r)$$

where the probability distributions $b_r(x_r), b_p(x_p)$ are proportional to $\exp(\hat{\phi}(x_r)/c_r)$ and $\exp(\hat{\phi}(x_p)/c_p)$ respectively while $\hat{\phi}$ is defined in Theorem 1. Whenever $c_r$ or $c_p$ are zero, their respective (sub)gradients are their max-beliefs or equivalently probability distributions over their maximal arguments of $\hat{\phi}$, c.f., Danskin theorem [1].

The optimal dual variables are those for which $\partial D / \partial \lambda_{r \to p}(x_r) = 0$, i.e., the corresponding beliefs agree on their marginal probabilities. When setting $\mu_{p \to r}(x_r)$ as above, the marginalization of $b_p(x_p)$ equals

$$\sum_{x_p \backslash x_r} b_p(x_p) \propto \exp\Big((\mu_{p \to r}(x_r) + \lambda_{r \to p}(x_r))/c_p\Big)$$

Therefore, by taking the logarithm, the gradient vanishes whenever the beliefs numerators agree up to an additive constant

$$\frac{\mu_{p \to r}(x_r) + \lambda_{r \to p}(x_r)}{c_p} = \frac{\phi'_r(x_r) - \sum_{p \in P(r)} \lambda_{r \to p}(x_r)}{c_r} \quad (3)$$

where $\phi'_r(x_r) = \phi_r(x_r) + \sum_{c \in C(r)} \lambda_{c \to r}(x_c)$. Multiplying both sides by $c_r c_p$ and summing both sides with respect to $p' \in P(r)$ we are able to obtain

$$\sum_{p' \in P(r)} \lambda_{r \to p'}(x_r) = c_{p,r} \Big( \phi'_r(x_r) + \sum_{p' \in P(r)} \mu_{p' \to r}(x_r)) \Big)$$

Plugin it into equation (3) results in the desired block dual descent update rule, i.e., $\lambda_{r \to p}(x_r)$ for which the partial derivatives vanish. □

For convenience, the explicit update rules also appear in Fig. 1. The above theorem provides a closed-form solution for block coordinate descent on the dual objective. Since the dual values are always lower bounded by the primal values, this algorithm is guaranteed to converge. However, the dual objective is not strictly convex therefore, a-priori, coordinate descent might not converge to the optimal solution [27]. Nevertheless, our proof technique demonstrates the dual-primal relation of dual coordinate descent, as throughout its runtime it generates a sequence of primal variables, namely the beliefs, which agree on their marginal probabilities. In the following we describe the conditions for which we can recover the primal optimal and dual optimal solutions.

**Theorem 3.** *The dual block descent algorithm in Theorem 2 is guaranteed to converge whenever $c_r$ are non-negative. Moreover, if $c_r > 0$ the dual block descent algorithm converges to the dual optimal value, and the beliefs $b_r(x_r)$ that are generated throughout the algorithm runtime converge to the primal optimal solution ; whenever its dual sequence is bounded every of its limit points is an optimal dual solution.*

**Proof:** See supplementary material. □

The above theorem does not constrain the dual variables, namely the messages. Specifically, although the dual value is guaranteed to converge to the optimum, the messages are not guaranteed to be bounded. This may happen since the dual function is not strictly convex, and the messages can be updated along an unbounded plateau that does not reduce the dual objective. The above result can be extended to guarantee optimality even if $c_r > 0$ only for the maximal regions, i.e., regions that are not contained by other regions. However, this result is mathematically involved as it does not correspond anymore to Bregman divergences and it is beyond the scope of this work.

The algorithm in Theorem 2 has an important meaning even when the primal function is not concave. In this case we lose all convergence guarantees, but the algorithm represents the Lagrangian saddle points. Therefore, whenever the algorithm converges it reaches a local maximum of the primal program.

**Theorem 4.** *Assume the program in equation (2) has mixed numbers $c_r \gtrless 0$. Then this program is not concave and the algorithm in Theorem 2 is not guaranteed to converge, but whenever it converges it reaches a stationary point for this program.*

**Proof:** See Supplementary material. □

The saddle point theorem does not consider programs for which the covering numbers can be zero, since in these cases we cannot uniquely restrict the beliefs. Nevertheless, this theorem applies to many important cases, such as the Bethe entropy and the tree reweighed entropies. In these cases, the negative coefficients for variables $c_i = 1 - \sum_{\alpha \supset i} c_\alpha$, where $\alpha$ are the non-singleton regions, play an important role. Specifically, the Bethe entropy provides an exact characterization for the entropy function over tree models, and the tree-reweighted entropy provides tighter bounds than the fractional covering bounds when restricting the fractional coverings to pairwise entropies. Unfortunately, the message-passing algorithm is not guaranteed to converge in these cases but the theorem above proves that when it converges it reaches a (local) maximum of the model.

## 4 Tightening High-Order Upper Bounds

Up to this point, the fractional covering numbers $c_r$ were held fixed, while we computed the upper bound for the partition function that is described in equation (2). We now consider how to find the optimal covering numbers which minimize the upper bound, while assuming we are able to compute the upper bound and its optimal arguments efficiently as described in Fig. 1. First we state some of the properties of these upper bounds as a function of their fractional covering numbers:

**Theorem 5. (Danskin Theorem)** *Denote the log-partition upper bound by*

$$U(c) = \max_{b_r(x_r) \in L(G)} \left\{ \sum_{r,x_r} b_r(x_r)\phi_r(x_r) + \sum_r c_r H(b_r) \right\}$$

*Assume $c_r \geq 0$ and that the maximal regions covering numbers are positive. Then $U(c)$ is a convex and differentiable function, and its partial derivatives are*

$$\frac{\partial U}{\partial c_r} = H(b_r^*)$$

*where*

$$b_r^*(x_r) = \operatorname*{argmax}_{b_r(x_r) \in L(G)} \left\{ \sum_{r,x_r} b_r(x_r)\phi_r(x_r) + \sum_r c_r H(b_r) \right\}$$

**Proof:** This is a special case of Danskin theorem, since $L(G)$ is a compact set. [1] □

To find the minimal upper bound on the partition function we need to minimize a convex and differentiable function over the convex set of fractional covering.

$$\min_{c_r \geq 0, \sum_{r \in R_i} c_r = 1} U(c)$$

Minimizing a convex function over a convex set is typically done using the conditional gradient method, for which one finds a direction of descent while respecting the fractional covering constraints [2]. Finding the direction of descent amounts to solving the linear program

$$d = \operatorname*{argmin}_{c_r \geq 0, \sum_{r \in R_i} c_r = 1} \sum_r c_r H(b_r^*) \quad (4)$$

The direction of descent is always attained, since it is the minimal argument of a continuous function over a compact set. However, when considering

> **Tightening the fractional covering upper bound:** Given functions $\phi_i(r_i)$. For every initial values of $\lambda_i(r)$:
>
> 1. For $t = 1, 2, ...$
>
>    (a) For $r \in \mathcal{R}$ do:
>    $$\forall i \in r \quad \lambda_i(r) = \epsilon \log \Big( \sum_{r_i \neq r} \exp \big( (\phi_i(r_i) + \lambda_i(r_i)) \big/ \epsilon \big) \Big) - \phi_i(r)$$
>    $$\text{additively normalize } \lambda_i(r) \text{ such that } \sum_{i \in r} \lambda_i(r) = 0$$
>
> 2. Output: $\forall i$, set $q_i(r_i) \propto \exp \big( (\phi_i(r_i) + \lambda_i(r_i)) \big/ \epsilon \big)$ and $\forall r$ set $c_r = q_i(r)$, for any of $i \in r$.

Figure 2: *The dual decomposition algorithm for recovering the direction of descent, described in equation (4). This algorithm can be seen as message-passing by reformulating the indexes $\lambda_i(r) \leftrightarrow \lambda_{i \to r}$. This demonstrates the differences between our two algorithms. The similarity between the two algorithms is a consequence of the block dual steps and the use of the entropy barrier method.*

large-scale problems, this linear program cannot be efficiently solved using standard linear programming solvers. For example, the simplex algorithms or interior point methods typically involve inverting the constraints matrix, an operation that cannot be done efficiently when dealing with graphical models that consist of millions of regions.

In this section we develop a new type of efficient linear programming solver for finding the minimal upper bound. Importantly, we cannot use the message-passing solvers in Section 3 to minimize these upper bounds, since the objective as well as the constraint sets in equation (4) are significantly different. However, we are able to construct efficient solvers while using the entropy barrier function in a similar manner. This emphasizes the computational importance of the entropy barrier function, since it results in closed-form update rules which turn to be crucial for large-scale linear programs.

The constraints of the linear program in equation (4) restrict the covering numbers $(c_r)_{r \in R_i}$ to the probability simplex. In order to use dual decomposition effectively, we need to decouple these simplex constraints, such that they restrict *separate* distributions. Therefore we inflate the covering numbers $c_r$ to non-overlapping probability distribution $q_i(r_i)$ for every $i = 1, ..., n$, while enforcing these distributions to agree, namely $q_i(r) = q_r$ for every $i \in r$. One can verify that these constraints are equivalent to the constraints over $c_r$ in equation (4). Thus we reformulate the linear program in equation (4) as

$$\underset{\substack{\forall i\ q_i\ \text{is probability} \\ \forall i \in r\ q_i(r) = q_r}}{\arg\min} \sum_{i, r_i} q_i(r_i) \Big( H(b^*_{r_i})/|r_i| \Big)$$

For computational efficiency we solve the above linear program by dual block ascent. Unfortunately, dual coordinate ascent may be sub-optimal when the dual program is not smooth, or equivalently the primal program is not strictly convex. Therefore we use the entropy barrier method to make the primal program strictly convex, which results in a smooth dual program. The following theorem shows that using the barrier method does not affect the quality of the solution.

**Theorem 6.** *Let $\phi_i(r_i) = H(b^*_{r_i})/|r_i|$ and consider the linear program for finding the direction of descent*

$$\underset{\substack{\forall i\ q_i\ \text{is probability} \\ \forall i \in r\ q_i(r) = q_r}}{\min} \sum_{i, r_i} q_i(r_i) \phi_i(r_i)$$

*Then the primal-dual programs*

*(primal)*
$$\underset{\substack{\forall i\ q_i\ \text{is probability} \\ \forall i \in r\ q_i(r) = q_r}}{\min} \sum_{i, r_i} q_i(r_i) \phi_i(r_i) - \epsilon \sum_i H(q_i)$$

*(dual)*
$$\underset{\sum_{i \in r} \lambda_i(r) = 0}{\max} \epsilon \sum_i \log \Big( \sum_{r_i} \exp \big( (\phi_i(r_i) + \lambda_i(r_i)) \big/ \epsilon \big) \Big)$$

*are $\delta-$approximations of the original linear program, where $\delta = \sum_i \epsilon \log |r_i|$*

**Proof:** The entropy $H(q_i)$ is a non-negative measure, thus the primal program lower bounds the original linear program. The bound holds since $H(q_i) \leq \log |r_i|$. The dual program is also a $\delta-$approximation since strong duality holds. ∎

Having a smooth dual program which relates to a strictly convex primal program, we can perform dual block ascent to reach their optimum. Using the entropy barrier function enables us to derive a closed-form update rules.

**Theorem 7.** *Consider the primal and dual programs in Theorem 6. Then the block dual ascent takes the form*

$$\forall r, \ \forall i \in r$$
$$\lambda_i(r) = \epsilon \log \Big( \sum_{r_i \neq r} \exp \Big( (\phi_i(r_i) + \lambda_i(r_i))\Big/\epsilon \Big) \Big) - \phi_i(r)$$

*additively normalize* $\lambda_i(r)$ *such that* $\sum_{i \in r} \lambda_i(r) = 0$

*Also, block dual ascent is guaranteed to converge to the dual optimum, the probabilities* $q_i(r_i)$ *that are generated throughout the algorithm runtime converge to the primal optimal point and whenever the dual sequence is bounded it is guaranteed to converge to an optimal dual solution.*

**Proof:** See supplementary material. □

For convenience, the algorithm appears in its explicit form in Fig. 2. The above block dual ascent algorithm can be seen as message-passing, where messages are sent between regions and variables, $\lambda_i(r) \leftrightarrow \lambda_{i \to r}$. This demonstrates the differences between our two algorithms, where the first sends two types of real-valued vectors $\lambda_{c \to p}(x_c), \mu_{p \to c}(x_c)$ along the edges of the regions graph, and the second sends a real number $\lambda_{i \to r}$ between variables and regions. The similarity between the two algorithms is a consequence of the block dual steps and the entropy barrier function.

## 5 Empirical Evaluation

In our experiments we first compared our message-passing algorithm for computing the fractional covering upper bound in equation (2) with the current state-of-the-art solver [12]. We compared these algorithms on the grid shape spin glass model. A spin glass model consists of $n$ spins $x_1, ..., x_n \in \{-1, 1\}$, whose local potentials are $\phi_i(x_i) = \theta_i x_i$ and its pairwise potentials are $\phi_{i,j}(x_i, x_j) = \theta_{i,j} x_i x_j$. The field parameters $\theta_i$ were chosen uniformly at random from $[-0.05, 0.05]$ and the coupling parameters $\theta_{i,j}$ were chosen uniformly at random from $[-1, 1]$. We used the same covering numbers for all edges and squares in the grid, setting $c_{i,j} = c_{i,j,k,l} = 1/9$ and $c_i = 1 - \sum_{r \in R_i \setminus i} c_r$. Therefore we satisfy the covering number constraints, $c_r \geq 0$ and $\sum_{r \in R_i} c_r = 1$. Our algorithm described in Fig. 1 used a region graph which connects squares to pairs and pairs to singletons, namely the Hesse diagram. [12] use a bipartite inner-outer region graph, for which all outer regions, i.e., squares, are connected to all inner regions, i.e., pairs and singletons. We used the same Matlab code, on a single core of Intel I5 with 8GB RAM, for both algorithm as our algorithm can be applied to bipartite region graphs as well. We compared both methods on $5 \times 5$ ,..., $100 \times 100$ grids and the stopping criteria was a primal-dual gap of $10^{-4}$. Fig. 3 shows that passing messages over the Hesse-diagram is better than working over the bipartite inner-outer region graph, and the gap between these methods is significant for large graphical models. We attribute this behavior to the fact that on the Hesse diagram we can pass many messages between pairs and singletons, while using time consuming square messages only when they are needed. In contrast [12] use square messages in every message-passing iteration.

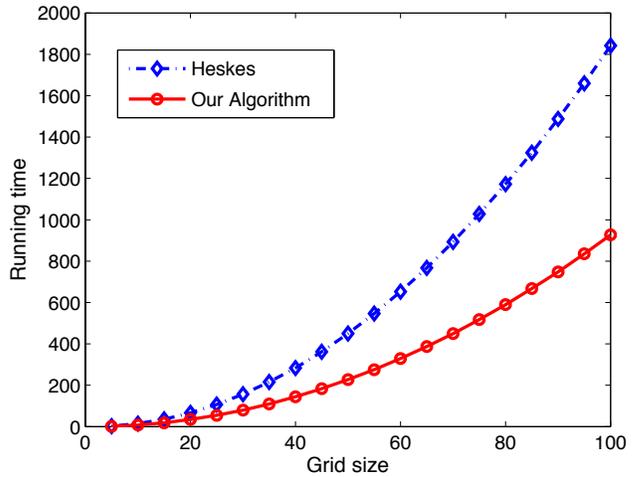

Figure 3: *Comparing our message-passing algorithm for computing the fractional covering upper bound in Section 3 with [12]. Comparison is for $5 \times 5$,..., $100 \times 100$ grid shape spin glass models, with singletons, pairs and squares regions. Our approach can utilize intermediate size message, between pairs and singletons, while [12] use the inner-outer region graph, thus sending square based messages in every iteration.*

In our experiments we also compared our dual decomposition algorithm for recovering the direction of descent over the fractional covering numbers in the tightening procedure, as described in Fig. 2. In this experiment we used the singletons, pairs and squares regions that correspond to the grid shape graphs. We used $\phi_i(r_i) = H(b_r^*)/|r|$, where $b_r^*(x_r)$ were the optimal beliefs in our previous experiments. The results in Fig. 4 show that the state-of-the-art off-the-shelf solver, the CPLEX, is good for small scale problems but significantly worse when applied to large scale graphical models. We note that our implementation is in Matlab, which has a significant overhead when applied to small scale problems.

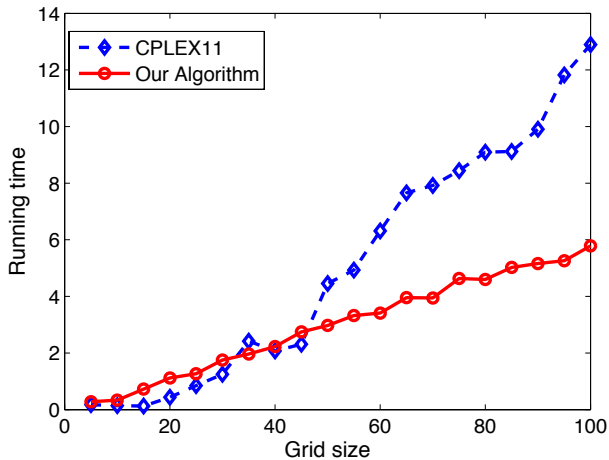
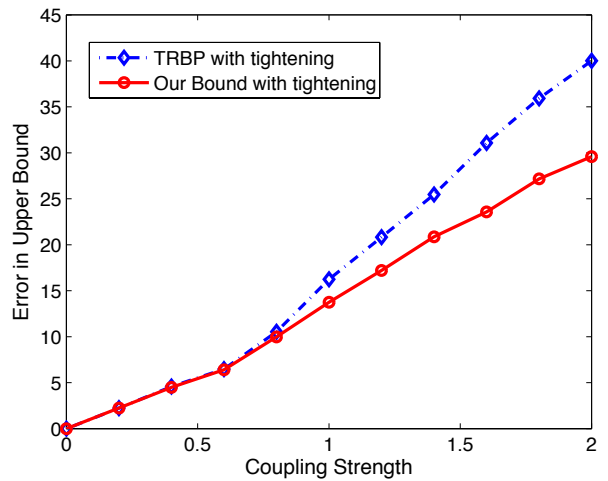

Figure 4: *Comparing our dual decomposition algorithm for recovering the direction of descent over fractional covering numbers in Section 4 with the CPLEX algorithm. Comparison is for singletons, pairs and squares regions that correspond to $5 \times 5,..., 100 \times 100$ grid shape graphs.*

Figure 5: *Comparing tightening tree-reweighted bounds with fractional covering bounds over grid shaped graphs. Our approach can use covering numbers of high-order regions therefore achieves better bounds.*

Tightening the fractional covering upper bounds can be applied to general region graphs. However, we cannot compare our approach on high-order region graph, as all previous approaches are efficient only when the regions consist of at most pairs of indexes. Therefore we compared our approach to tightening tree reweighed upper bounds [32] implemented by [14]. In our experiments we used the grid shaped spin glass model with local field parameters $\theta_i \in [-0.05, 0.05]$ and mixed coupling potentials $\theta_{i,j} \in [-c, c]$ that ranges over $c = 0, ..., 2$. Our tightening algorithm used singletons, pairs and squares regions. The results appear in Fig. 5 showing that going to high-order regions graphs provides tighter upper bounds.

## 6 Related Work

In this work we investigate upper bounds on the partition function over regions graphs. For this purpose we use the known fractional covering upper bounds for the entropy function [6, 23]. We applied these bound to the partition function through conjugate duality.

Tightening upper bounds for the partition function results in two programs: A concave program for computing the upper bounds and a convex program which tightens these bounds. The field of convex (or concave) optimization is large and contains many different solvers. However, most of these solvers cannot be applied efficiently to our problems, since they ignore their structures [3, 1]. Our work differs from these works in an important respect, as we exploit the structure of the problem, decomposing the constraints through the dual program and then performing efficient closed-form steps in every iteration. This block coordinate steps in the dual enable us to efficiently deal with large-scale problems.

Upper bounds on the partition function were extensively studied in the last decade. [32] presents upper bounds for graphical models that are based on spanning trees, as well as a method to tighten these bounds using conditional gradient descent. The conditional gradient is recovered by looking for a maximal spanning tree in the graphical model. The upper bound for the graphical model is computed by a message-passing algorithm called sum-TRBP, a method that extends to region graphs [35]. This work differs from ours in important respects: First, the spanning tree method best fits graph and encounters computational difficulties when dealing with hypergraphs, or equivalently region graphs. Working with region graphs, their conditional gradient method looks for a spanning hypertree, which is a NP-hard problem [17]. This problem was already pointed out in [32] and motivates this work. Thus in our work we suggest to use the known fractional covering upper bounds over regions graphs [6, 23], and we compute the conditional gradient through dual decomposition and the entropy barrier method. Second, the sum-TRBP and its high-order extension compute a bound that has covering numbers with mixed signs. Thus the sum-TRBP is not guaranteed to converge. In contrast our work considers bounds that have only positive covering numbers, thus our bound computation is guaranteed to converge to the optimum. We note that there are other algorithms that fix the convergence of sum-TRBP, but these algorithms cannot be applied to high-order regions graphs [9, 11].

Convexity provides a well established framework to extend the spanning trees upper bounds to other combinatorial objects. [8] describe upper bounds based on planar decomposition. Since planarity is a property of pairwise regions this work does not extend to general region graphs. In contrast, our work presents efficient bounds and tightening for high-order region graphs. [7] provide a family of upper bounds that are based on conditional entropy decompositions. These upper bounds can be applied to high-order region graphs. However, this work differs from ours in important respects: First, to compute the bound they directly solved the primal program. Since the primal program consists of conditional entropies, which are not strictly concave, they used a conditional gradient solver which is suboptimal when addressing large scale region graphs. A subsequent work presented the corresponding message-passing solver, but it was restricted to pairwise regions [9]. In contrast, our work introduces the fractional covering entropy bounds which are strictly concave, thus providing an efficient dual decomposition solver through message-passing over the region graph. Second, their tightening approach considers all conditional entropies sequences, thus they are restricted to a small number of sequences. In contrast, our work provides an efficient dual decomposition algorithm to tighten the bound over all fractional covering numbers. [21] introduce a new approach for conditional entropy decomposition, which allows to use more sequences through a mini-bucket elimination method. However, this approach suffers from similar drawbacks as [7] when applied to region graphs.

In Fig. 1 we present a message-passing algorithm to compute upper bounds on the partition function in high-order graphical models. This algorithm can be applied to general programs, consisting of sums of linear terms and entropy terms. In the last decade many different message-passing algorithms were devised for similar programs. Assuming all regions intersect on at most one variable, our algorithm has two types of messages, $\lambda_{i \to \alpha}(x_i)$ and $\mu_{\alpha \to i}(x_i)$, and it reduces to the norm-product belief propagation, which includes as special cases both sum-product and max-product, tree-reweighted belief propagation, NMPLP and the asynchronous splitting algorithm for different values of $c_i, c_\alpha$ [26, 32, 33, 10, 28]. However, the main purpose of our algorithm is high-order region graphs. In this perspective, when the region graph is a bipartite graph that contains outer-inner regions, our algorithm reduces the parent-to-child generalized belief propagation algorithm [36, 16, 13], and its convex forms [12, 24]. These works differ from ours in an important respect, as they consider a bipartite region graph which is computationally demanding since it uses the maximal regions in every iterations. In contrast, our algorithm considers a general region graph, thus enables to pass messages between intermediate size regions to gain computational efficiency.

# 7 Conclusions and Discussion

In our work we describe methods to tighten upper bounds on the partition function over general region graphs. These upper bounds use the fractional covering bounds on the entropy function. We introduced two dual decomposition algorithms, for computing the upper bound and to minimize the upper bound. Both algorithms use the entropy barrier function to obtain a closed-form block dual steps that optimize their respective dual programs.

To compute the upper bound we solve a concave program, consisting of linear terms and entropy terms. Our solver passes messages along the edges of the region graph that describe these terms. The computational complexity of this solver depends on the structure of the region graph. It turns out that there are many different region graphs, such as the inner-outer graph or the Hesse diagram that describe the same program [12, 36]. Although some works reasoned about the optimal graph, this problem is largely open [25].

Interestingly, the message-passing algorithm for computing the upper bound sends partial log-partition functions $\mu_{p \to c}(x_c)$ weighted by the covering number $c_r$. In particular, when $c_r = 1$ it sends a sum-product based message and when $c_r = 0$ is sends a max-product based messages. The form of these messages is determined by the covering number, and their primal interpretation relates to the weight $c_r$ of the corresponding entropy function. For example, whenever no entropy terms are used, i.e., we use the algorithm for solving linear program relaxations, we only use max-product based operations and our algorithm contains the max-product, max-TRBP, NMPLP and convex max-product as special cases. In this perspective we could have used this algorithm to tighten linear program relaxations, but this problem was solved by [29]. Surprisingly, using fractional covering upper bounds while some of the covering numbers equal zero, we get a mixture of sum-product and max-product rules. This is a result of having sum of entropy and non-entropy terms in the primal. This goes against the common practice that mixing max-product and sum-product rules relates to the marginal-MAP solution [18, 22, 15]. The relations and differences between these two problems are subject to further research.